\documentclass[fleqn,10pt]{wlscirep}
\usepackage[utf8]{inputenc}
\usepackage[T1]{fontenc}
\usepackage{breakurl}
\usepackage{url}
\usepackage{multirow}  
\usepackage{makecell}
\usepackage{threeparttable}  

\title{CMR$\times$Recon: An open cardiac MRI dataset for the competition of accelerated image reconstruction}

\author[1,$\#$]{Chengyan Wang}
\author[2,$\#$,$\ddag$]{Jun Lyu}
\author[3,+]{Shuo Wang}
\author[4,$\dag$]{Chen Qin}
\author[5]{Kunyuan Guo}
\author[1]{Xinyu Zhang}
\author[5]{Xiaotong Yu}
\author[6]{Yan Li}
\author[7]{Fanwen Wang}
\author[8]{Jianhua Jin}
\author[9]{Zhang Shi}
\author[10]{Ziqiang Xu}
\author[11]{Yapeng Tian}
\author[12]{Sha Hua}
\author[13]{Zhensen Chen}
\author[1]{Meng Liu}
\author[1]{Mengting Sun}
\author[1]{Xutong Kuang}
\author[3]{Kang Wang}
\author[3]{Haoran Wang}
\author[13]{Hao Li}
\author[14]{Yinghua Chu}
\author[7]{Guang Yang}
\author[15,16]{Wenjia Bai}
\author[8]{Xiahai Zhuang}
\author[1,13]{He Wang}
\author[2,*]{Jing Qin}
\author[5,*]{Xiaobo Qu}

\affil[1]{Human Phenome Institute, Fudan University, Shanghai, China}
\affil[2]{School of Nursing, The Hong Kong Polytechnic University, Hong Kong, China}
\affil[3]{Digital Medical Research Center, School of Basic Medical Sciences, Fudan University, Shanghai, China}
\affil[4]{Department of Electrical and Electronic Engineering \& I-X, Imperial College London, United Kingdom}
\affil[5]{Department of Electronic Science, Fujian Provincial Key Laboratory of Plasma and Magnetic Resonance, National Institute for Data Science in Health and Medicine, Institute of Artificial Intelligence, Xiamen University, Xiamen, China}
\affil[6]{Department of Radiology, Ruijin Hospital, Shanghai Jiao Tong University School of Medicine, Shanghai, China}
\affil[7]{Department of Bioengineering\~/Imperial-X, Imperial College London, United Kingdom }
\affil[8]{School of Data Science, Fudan University, Shanghai, China}
\affil[9]{Department of Radiology, Zhongshan Hospital, Fudan University, Shanghai, China}
\affil[10]{School of Health Science and Engineering, University of Shanghai for Science and Technology, Shanghai, China}
\affil[11]{Department of Computer Science, The University of Texas at Dallas, United States}
\affil[12]{Department of Cardiovascular Medicine, Ruijin Hospital Lu Wan Branch, Shanghai Jiao Tong University School of Medicine, Shanghai, China}
\affil[13]{Institute of Science and Technology for Brain-Inspired Intelligence, Fudan University, Shanghai, China, 200433}
\affil[14]{Simens Healthineers Ltd., China}
\affil[15]{Department of Brain Sciences, Imperial College London, London, UK}
\affil[16]{Department of Computing, Imperial College London, London, UK}

\affil[$\#$]{\textbf{These authors contributed equally to this work: wangcy@fudan.edu.cn; junlyu@polyu.edu.hk}}
\affil[*]{\textbf{These authors are co-corresponding authors: harry.qin@polyu.edu.hk; quxiaobo@xmu.edu.cn}}
\affil[+]{\textbf{Image quality control contact: shuowang@fudan.edu.cn}}
\affil[$\dag$]{\textbf{Primary contact person of Task 1: c.qin15@imperial.ac.uk}}
\affil[$\ddag$]{\textbf{Primary contact person of Task 2: junlyu@polyu.edu.hk}}

\begin{abstract}
Cardiac magnetic resonance imaging (CMR) has emerged as a valuable diagnostic tool for cardiac diseases. However, a limitation of CMR is its slow imaging speed, which causes patient discomfort and introduces artifacts in the images. There has been growing interest in deep learning-based CMR imaging algorithms that can reconstruct high-quality images from highly under-sampled k-space data. However, the development of deep learning methods requires large training datasets, which have not been publicly available for CMR. To address this gap, we released a dataset that includes multi-contrast, multi-view, multi-slice and multi-coil CMR imaging data from 300 subjects. Imaging studies include cardiac cine and mapping sequences. Manual segmentations of the myocardium and chambers of all the subjects are also provided within the dataset. Scripts of state-of-the-art reconstruction algorithms were also provided as a point of reference. Our aim is to facilitate the advancement of state-of-the-art CMR image reconstruction by introducing standardized evaluation criteria and making the dataset freely accessible to the research community. Researchers can access the dataset at \url{https://www.synapse.org/\#!Synapse:syn51471091/wiki/}.
\end{abstract}
\begin{document}

\flushbottom
\maketitle
\thispagestyle{empty}


\section*{Background \& Summary}

Cardiac magnetic resonance (CMR) imaging has emerged as a valuable technique for diagnosing cardiovascular diseases, thanks to its superior soft tissue contrast and non-invasive nature. CMR imaging can provide anatomical, functional, and characteristic information about the heart, and has been reported in many population studies ~\cite{nakamura2018mechanisms, jaffe2006biomarkers,bai2020population,bai2021longitudinal,wang2021recommendation}. Cine magnetic resonance imaging (MRI) is currently considered the gold standard for non-invasive evaluation of cardiac functionality. Cardiac T1 and T2 quantitative mapping have been widely used in evaluating intracellular disturbances of cardiomyocytes ~\cite{wang2021recommendation}. Myocardial T1 characterization is valuable for detecting and assessing various cardiomyopathies, while T2 changes have been observed in edematous regions in patients with infarction, hemorrhage, graft rejection, or myocarditis. 

However, a drawback of CMR is its inherently slow imaging speed, which can cause patient discomfort and result in motion artifacts in images. Typically, cine is obtained using electrocardiography (ECG)-gated segmented gradient sequence, as acquiring the full k-space within one acquisition window is not feasible. Thus, the entire k-space is read out over multiple cardiac cycles. However, for patients with impaired breath-hold capacity or cardiac arrhythmia, image degradation will occur due to the long acquisition time and motions, which can influence further diagnosis. Accelerated cine imaging addresses these limitations by filling up the k-space of all temporal phases in fewer cardiac cycles and breath-holds. Sufficient acceleration also enables "real-time" imaging, substantially reducing artifacts associated with respiratory motion and arrhythmia. Similar to cine imaging, conventional T1 and T2 mapping are prone to failure due to decreased image quality when faced with highly under-sampled multi-contrast CMR images. Accelerated T1 and T2 mapping shortens the acquisition window, leading to a significant reduction of artifacts associated with respiratory motion and arrhythmia ~\cite{wang2019black}. Consequently, there has been a growing interest in accelerated CMR image reconstruction from highly under-sampled k-space data. 

So far, many artificial intelligence (AI)-based image reconstruction algorithms have shown great potential in improving imaging performance ~\cite{lyu2023region,qin2018convolutional,qin2021complementary,lv2020parallel,lyu2022dudocaf,wang2022extreme,hammernik2018learning,aggarwal2018modl,knoll2020fastmri,zhao2022fastmri+,tibrewala2023fastmri}. However, deep-learning-based methods require large quantities of raw k-space data for model training. The field of CMR imaging still lacks standardized and high-quality datasets that are publicly available. In addition, due to the absence of large public datasets, there is no common gold standard on which they can be properly compared. To date, NYU Langone Health has released `fastMRI’ dataset, containing multi-channel brain, knee and prostate MRI raw data. However, these datasets are inadequate for the 3D+1D (time domain) scenario in cardiac imaging. The limited availability of raw k-space datasets motivated the release of this `CMR$\times$Recon' dataset. The goal of establishing the `CMR$\times$Recon' dataset is to provide a benchmark dataset that enables the broad research community to promote advances in high-quality CMR imaging. 

In this paper, we describe the first release of CMR raw k-space data that includes multi-contrast, multi-view, and multi-channel cardiac imaging from 300 subjects. Imaging studies include cine and mapping sequences. In addition, we released processed CMR images in The Neuroimaging Informatics Technology Initiative (NIFTI) format and the corresponding scripts of \emph{state-of-the-art} parallel imaging. Manual segmentations of the myocardium and chambers of all the subjects are also provided within the dataset by an experienced radiologist. 

\section*{Methods}
\subsection*{2.1 Subject characteristics}

The study was approved by the local institutional review board. All participants were aware of the nature of the study and agreed to make their materials publicly available in anonymized form. Inclusion criteria were defined as: 1) adults without a pathologically confirmed diagnosis of cardiovascular disease, and 2) availability of an MRI examination with all imaging sequences. A total of 300 healthy volunteers (160 females and 140 males) were recruited between June 2022 and March 2023 with written informed consent. The mean age of the subjects was 26 ± 5 years. 

\subsection*{2.2 Image acquisition}

\begin{figure}[t!]
\centering
\includegraphics[width=0.8\linewidth]{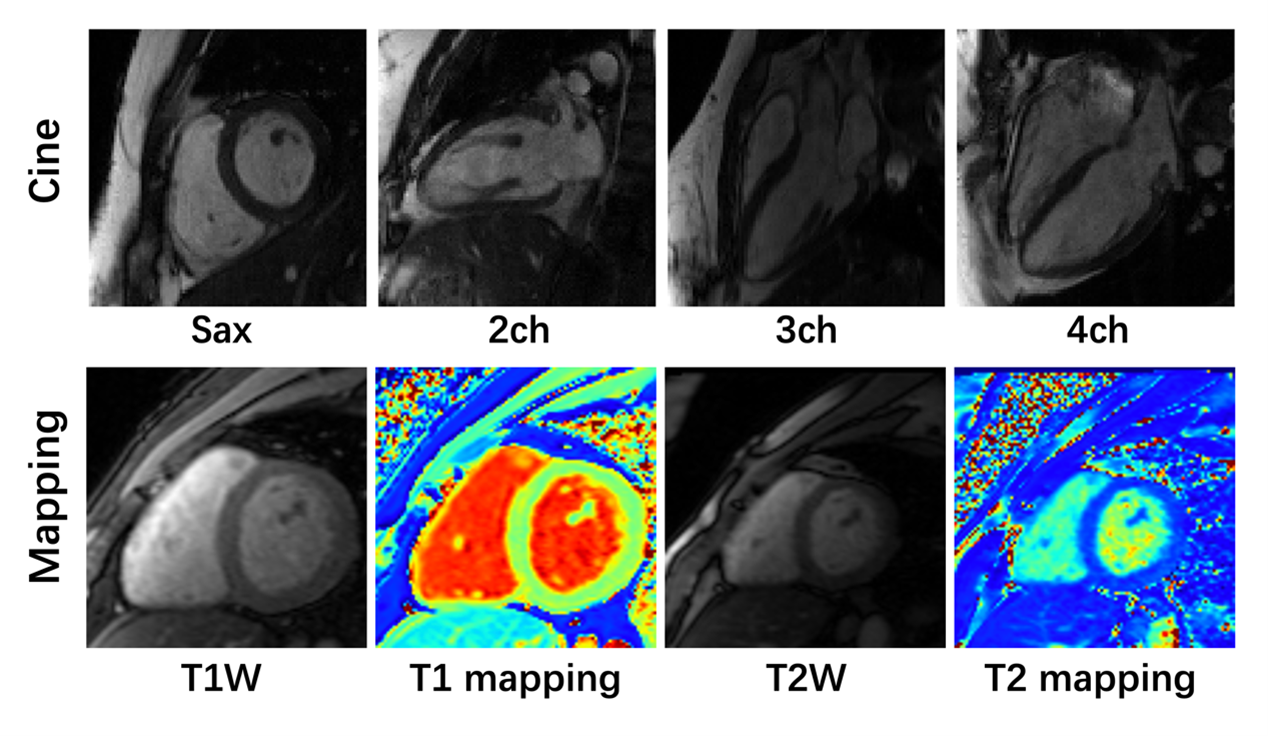}
\caption{Representative CMR images of cardiac cine and mapping. 
Note: SAX, T1W, T2W and ch means short-axis, T1 weighted, T2 weighted and chamber, respectively.
}
\label{fig:CMRI1}
\end{figure}

Data were acquired on a 3T scanner (MAGNETOM Vida, Siemens Healthineers, Germany), with a dedicated cardiac coil made up of 32 channels. Participants were placed in a supine position on the table before scans. Electrodes were attached and Electrocardiogram (ECG) signals were recorded during the scan. The `Dot’ engine was used for cardiac scout imaging. Figure ~\ref{fig:CMRI1} shows representative CMR images of cardiac cine and mapping.

We followed the recommendation of CMR imaging reported in the previous publication ~\cite{wang2021recommendation}. The `TrueFISP’ readout was used for 2D cardiac cine acquisitions. The collected images included short-axis (SAX), two-chamber (2CH), three-chamber (3CH) and four-chamber (4CH) long-axis (LAX) views. Typically, 5$\sim$11 slices were acquired for SAX view, while a single slice was acquired for each of the other views. The cardiac cycle was segmented into 12$\sim$25 phases with a temporal resolution $\sim$50 ms according to the heart rate. Typical scan parameters were: spatial resolution of 1.5$\times$1.5 mm$^2$, slice thickness of 8.0 mm, repetition time (TR) of 3.6 ms, and echo-time (TE) of 1.6 ms. The parallel imaging acceleration factor was $R$=3. Signals were acquired with breath-hold. The complete imaging parameters are summarized in Table~\ref{tab:tab1}.

T1 mapping was conducted using a modified look-locker inversion recovery (MOLLI) sequence, which acquired 9 images with different T1 weightings (using the 4-(1)-3-(1)-2 scheme). T1 mapping was performed in SAX view only, with typical field-of-view (FOV) of 340$\times$340 mm$^2$, spatial resolution of 1.5$\times$1.5 mm$^2$, slice number of 5$\sim$6, slice thickness of 5.0 mm, TR of 2.7 ms, TE of 1.1 ms, partial Fourier of 6/8, and parallel imaging acceleration factor of $R$=2. The inversion time varied among subjects according to the real-time heart rate. Signals were collected at the end of the diastole with ECG triggering. 

T2 mapping was performed using a T2-prepared (T2prep)-FLASH sequence with three T2 weightings in SAX view, with identical geometrical parameters as used in T1 mapping. Typical imaging parameters were FOV of 340$\times$340 mm$^2$, spatial resolution of 1.5$\times$1.5 mm$^2$, slice number of 5$\sim$6, slice thickness of 5.0 mm, TR of 3.0 ms, TE of 1.3 ms, T2 preparation time of 0/35/55 ms, partial Fourier of 6/8, and parallel imaging acceleration factor of $R$=2. Signals were collected at the end of the diastole with ECG triggering. 

\subsection*{2.3 Image Processing}
\begin{figure}[t!]
\centering
\includegraphics[width=0.8\linewidth]{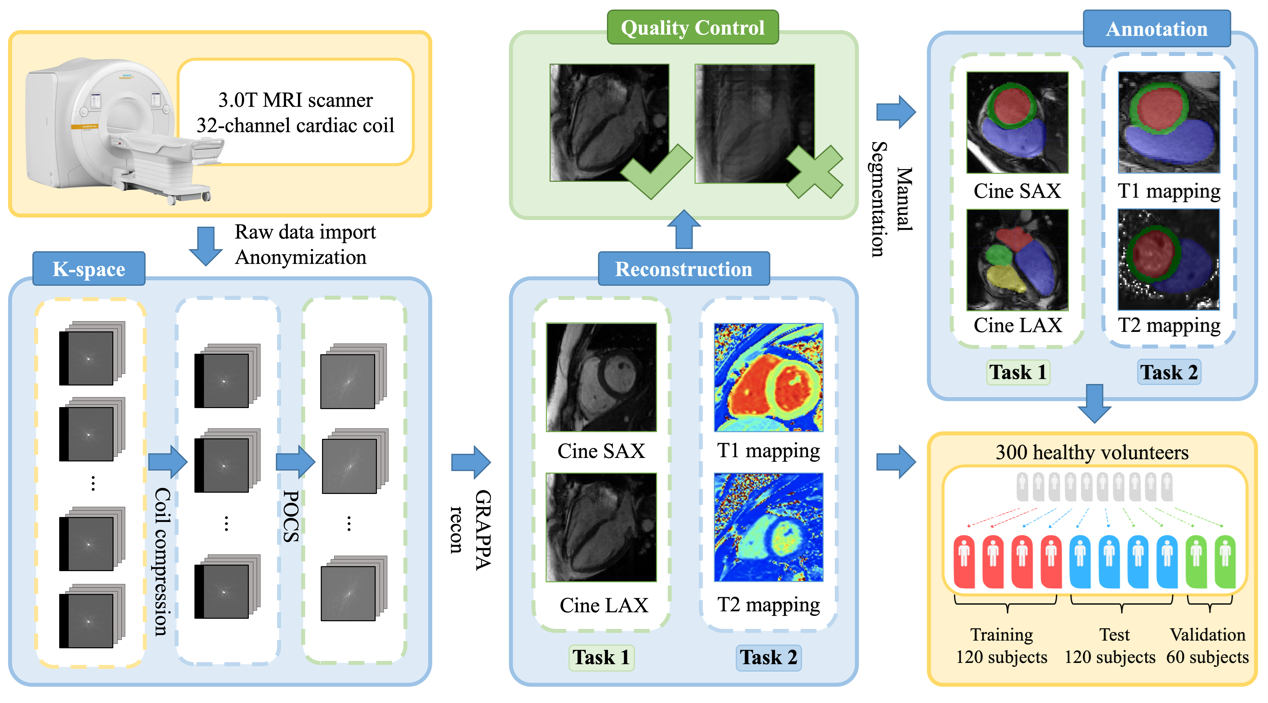}
\caption{General workflow to produce the `CMR$\times$Recon' dataset. Multi-contrast, multi-view, multi-slice, multi- channel k-space data were acquired from 300 healthy volunteers using a 3.0T MRI scanner equipped with a 32-channel cardiac coil. The final dataset includes a training set of 120 subjects, a validation subset of 60 subjects, and a test subset of 120 subjects.}
\label{fig:workflow2}
\end{figure}

The general workflow to produce the `CMR$\times$Recon' dataset is illustrated in Figure ~\ref{fig:workflow2}. The raw `.dat’ format data exported from the scanner were imported through the data interface provided by Siemens Healthineers. The k-space data were anonymized (including subject and center data anonymization) via conversion to the raw data format. The multi-coil k-space data were compressed to 10 virtual coils to reduce the burden of computation and storage ~\cite{zhang2013coil}. The partial Fourier data were filled up using projection onto convex sets (POCS) ~\cite{haacke1991fast} (integrated from code written by Michael Völker: michael.voelker@mr-bavaria.de). Each under-sampled k-space data was reconstructed using GeneRalized Autocalibrating Partially Parallel Acquisition (GRAPPA) ~\cite{zhao2022fastmri+} to acquire full k-space data. Those images with poor quality were removed based on visual assessment by experienced radiologists. After these processing steps, the resulting k-space was transformed to .mat format (MATLAB 2018a) while the corresponding images were stored in NITFI format. 

Training datasets include full k-space data, under-sampled k-space data with acceleration factors of 4, 8, and 10, sampling masks, auto-calibration (ACS) lines (24 lines) and reconstructed images. Validation data included under-sampled k-space data with acceleration factors of 4, 8, and 10, sampling mask, and ACS lines (24 lines). Test data include under-sampled k-space data with acceleration factors of 4, 8, and 10, sampling masks, and ACS lines (ACS, 24 lines). Undersampling was performed by retrospectively sub-sampling k-space lines from a full k-space in Cartesian trajectory (reconstructed by GRAPPA). k-space lines were omitted only in the phase encoding direction to simulate physically realizable accelerations in cine and mapping acquisitions. The same undersampling mask was applied to all slices in a volume. Uniform undersampling was performed for all acceleration factors. 

\subsection*{2.4 Manual segmentation}

\begin{figure}[ht]
\centering
\includegraphics[width=0.7\linewidth]{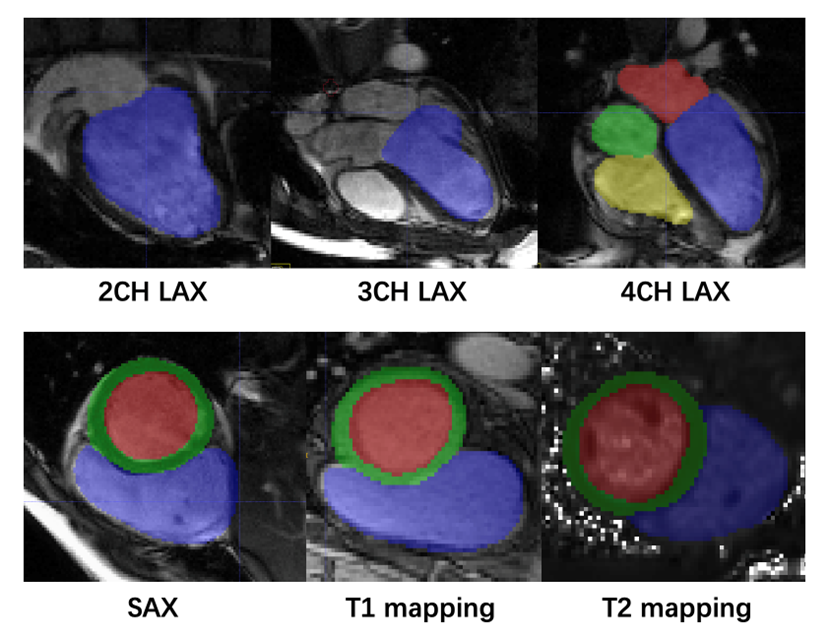}
\caption{Representative cine and mapping images with manually annotations.}
\label{fig:CMRI3}
\end{figure}

Manual segmentations of myocardium and chambers were performed by an experienced radiologist (with more than 5 years of cardiac imaging experience) using ITK-SNAP (version 3.8.0). The segmentation labels and the corresponding images were stored in NIFTI format, maintaining the original image coordinates. Representative images with manual annotations are shown in Figure ~\ref{fig:CMRI3}.

For the LAX cine images, four cardiac chambers have been labeled as follows: 
\begin{enumerate}
\item[a)] Left atrium - label 1; 
\item[b)] Right atrium - label 2; 
\item[c)] Left ventricle - label 3; 
\item[d)] Right ventricle - label 4. 
\end{enumerate}

For the SAX cine images, we performed the following labeling:
\begin{enumerate}
\item[a)] Left ventricle - label 1;
\item[b)] Left ventricular myocardium - label 2;
\item[c)] Right ventricle - label 3.
\end{enumerate}
The annotations of both T1 mapping and T2 mapping were the same as SAX cine. 

\subsection*{2.5 Parametric calculation}

A 3-parameter fitting model was used to calculate myocardial T1 values of each time course: 
\begin{center}
\begin{equation}
S(T I)=A-B e^{-T I / T_1^*}
\end{equation}
\end{center}
where S(TI) represents the signal intensity; TI represents the inversion time; A, B, and T1* are parameters to be estimated. In experiments, a total of 9 TIs (using the 4-(1)-3-(1)-2 scheme) were set to obtain the corresponding k-space for each subject. The inversion time varied among subjects according to the real-time heart rate.

T1 was determined from the resulting A, B, and T1* by applying the following equation:
\begin{center}
\begin{equation}
\mathrm{T}_1=\mathrm{T}_1^*(\mathrm{~B} / \mathrm{A}-1)
\end{equation}
\end{center}

The T2 value is calculated with equation 3: 
\begin{equation}
S(T E)=A \exp \left(-T E / \mathrm{T}_2\right)
\end{equation}
where S(TE) represents the signal intensity; TE represents the T2 preparation time; $T_2$ is the T2 value to be fitted.

It should be mentioned that we did not perform motion corrections before parametric fitting. Instead, we removed those images with unacceptable motion artifacts under the guidance of the radiologist. The quantitative fitting models and results may provide additional modules to enhance the CMR reconstruction performances. 

\section*{Data Records}
\subsection*{3.1 Data description}
The released dataset includes 120 training data, 60 validation data, and 120 test data. The training and validation datasets can be used to train reconstruction models and to determine hyperparameter values, while the test dataset is used to compare the results across different approaches. Table~\ref{tab:tab2} shows an overview of the `CMR$\times$Recon' dataset. 

\subsection*{3.2 Data format}
\begin{figure}[ht]
\centering
\includegraphics[width=0.9\linewidth]{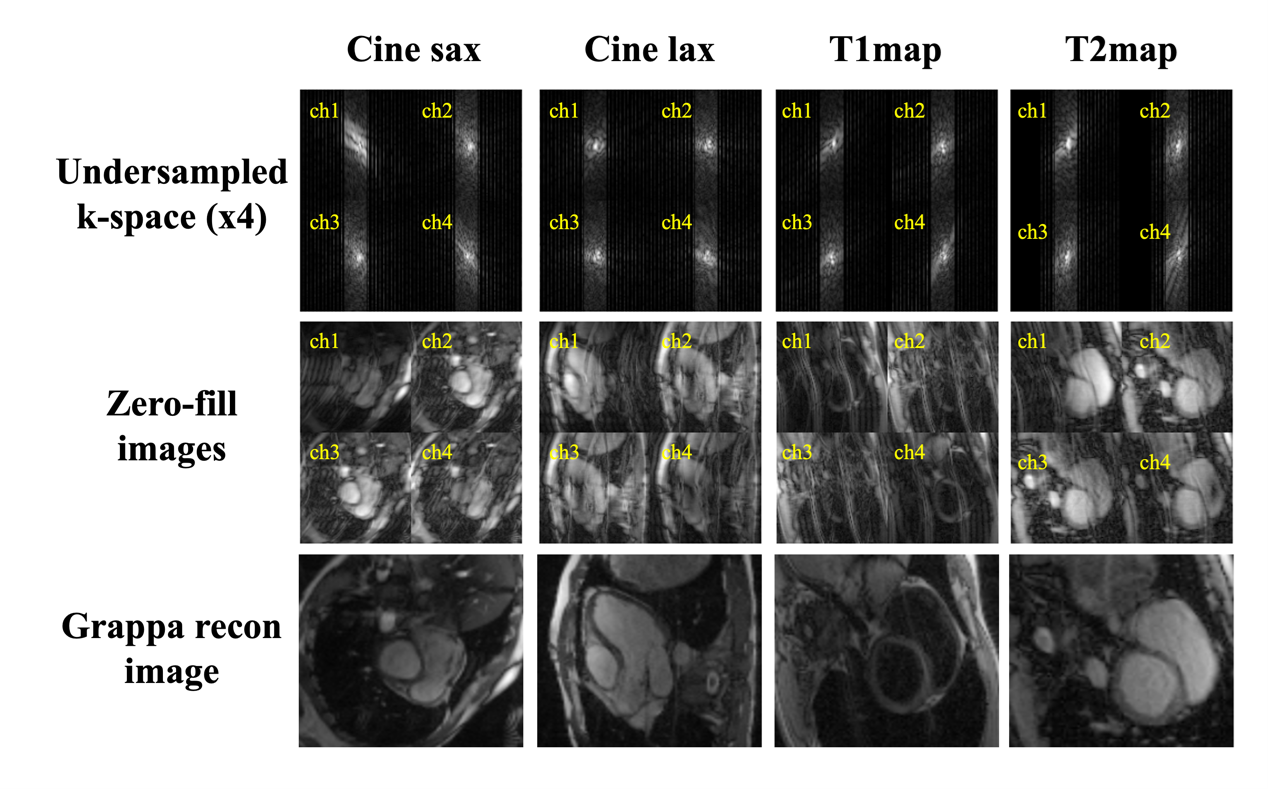}
\caption{Representative cardiac multi-coil data and reconstructed images.}
\label{fig:recI4}
\end{figure}

The CMR$\times$Recon dataset contains four types of k-space data, i.e., full-sampled data and 4$\times$, 8$\times$, 10$\times$ sub-sampled data, all of which were complex-valued single precision multi-coil data. Both multi-coil and emulated single-coil data were provided. We used the emulated single-coil methodology to simulate single-coil data from a multi-coil acquisition ~\cite{tygert2020simulating}. Detailed descriptions of the data types of cardiac cine and mapping are summarized in Table ~\ref{tab:tab3} and Table ~\ref{tab:tab4}. Representative multi-coil k-space data (with an acceleration factor of 4), zero-filling reconstructed images and GRAPPA reconstructed images are shown in Figure ~\ref{fig:recI4}.

\section*{Technical Validation}
\subsection*{4.1 Evaluation metrics }
The python scripts for image quality evaluation metrics were provided in Github: \url{https://github.com/CmrxRecon/CMRxRecon/tree/main/Evaluation}. 

For cardiac cine, the following criteria were used for reconstruction results assessments:
normalized mean square error (NMSE), peak signal-to-noise ratio (PSNR), and structural similarity index measure (SSIM). Definitions of the metrics were as follows:
\begin{enumerate}
\item[a)] NMSE

The NMSE between a reconstructed image or image volume represented as a vector $\hat{v}$ and a reference image or volume $v$ is defined as
\begin{equation}
\operatorname{NMSE}(\hat{v}, v)=\frac{\|\hat{v}-v\|_2^2}{\|v\|_2^2}
\end{equation}
where $\|\cdot\|_2^2$ is the squared Euclidean norm, and the subtraction is performed entry-wise. In this work, we report NMSE values computed and normalized over full image volumes rather than individual slices.

\item[b)] PSNR

The PSNR represents the ratio between the power of the maximum possible image intensity across a volume and the power of distorting noise and other errors:
\begin{equation}
\operatorname{PSNR}(\hat{v}, v)=10 \log _{10} \frac{\max (v)^2}{\operatorname{MSE}(\hat{v}, v)}
\end{equation}
Here $\hat{v}$ is the reconstructed volume, $v$ is the target volume, $max(v)$ is the largest entry in the target volume $v$, $\operatorname{MSE}(\hat{v}, v)$ is the mean square error between $\hat{v}$ and $v$ defined as $\frac{1}{n}\|\hat{v}-v\|_2^2$ and $n$ is the number of entries in the target volume $v$. 

\item[c)] SSIM

The SSIM index measures the similarity between two images by exploiting the inter-dependencies among nearby pixels. The resulting similarity between two image patches $\hat{m}$ and $m$ is defined as
\begin{equation}
\operatorname{SSIM}(\hat{m}, m)=\frac{\left(2 \mu_{\hat{m}} \mu_m+c_1\right)\left(2 \sigma_{\hat{m} m}+c_2\right)}{\left(\mu_{\hat{m}}^2+\mu_m^2+c_1\right)\left(\sigma_{\hat{m}}^2+\sigma_m^2+c_2\right)^{\prime}}
\end{equation}
where $\mu_{\hat{m}}$ and $\mu_{m}$ are the average pixel intensities in $\hat{m}$ and $m$, $\sigma_{\hat{m}}^2$ and $\sigma_{{m}}^2$ are their variances, $\sigma_{\hat{m}m}$ is the covariance between $\hat{m}$ and $m$, and $c_1$ and $c_2$ are two variables to stabilize the division; $c_1=(k_1 L)^2$ and $c_2=(k_2 L)^2$. 
\end{enumerate}

For T1 and T2 mapping, the root mean square error (RMSE) was calculated for reconstruction analysis.

\subsection*{4.2 Benchmark Reconstruction}

\begin{figure}[ht]
\centering
\includegraphics[width=0.9\linewidth]{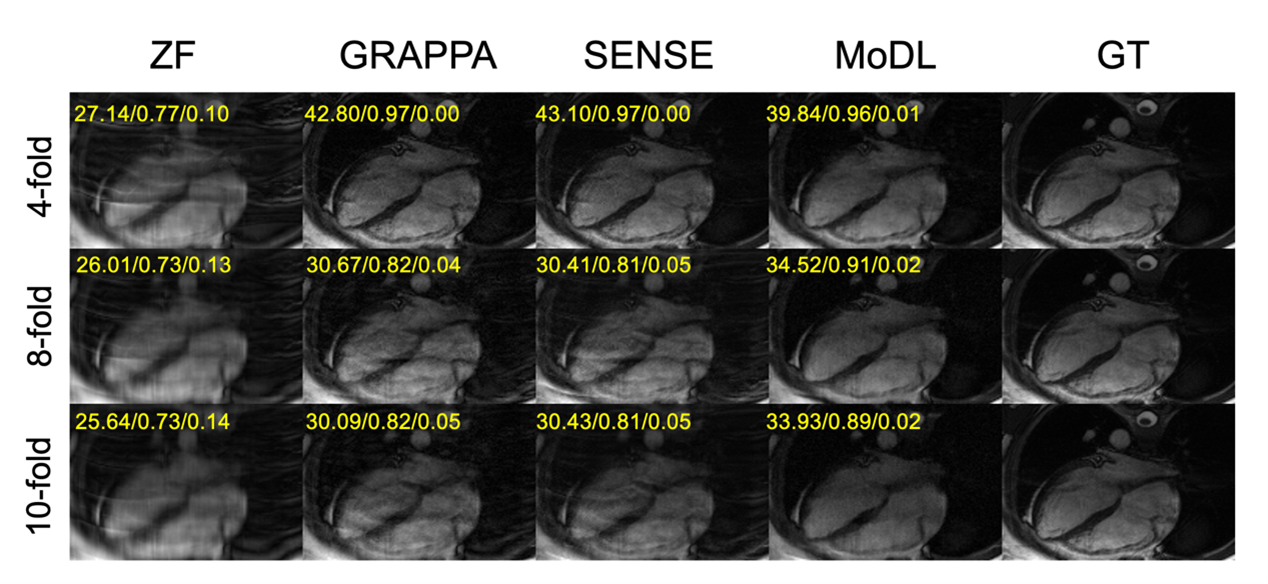}
\caption{Representative Cine long-axis images in the `CMR$\times$Recon' dataset reconstructed with benchmark reconstruction algorithms. The yellow numbers in each subgraph represent PSNR, SSIM and NMSE, respectively.}
\label{fig:recI5}
\end{figure}

\begin{figure}[t!]
\centering
\includegraphics[width=0.8\linewidth]{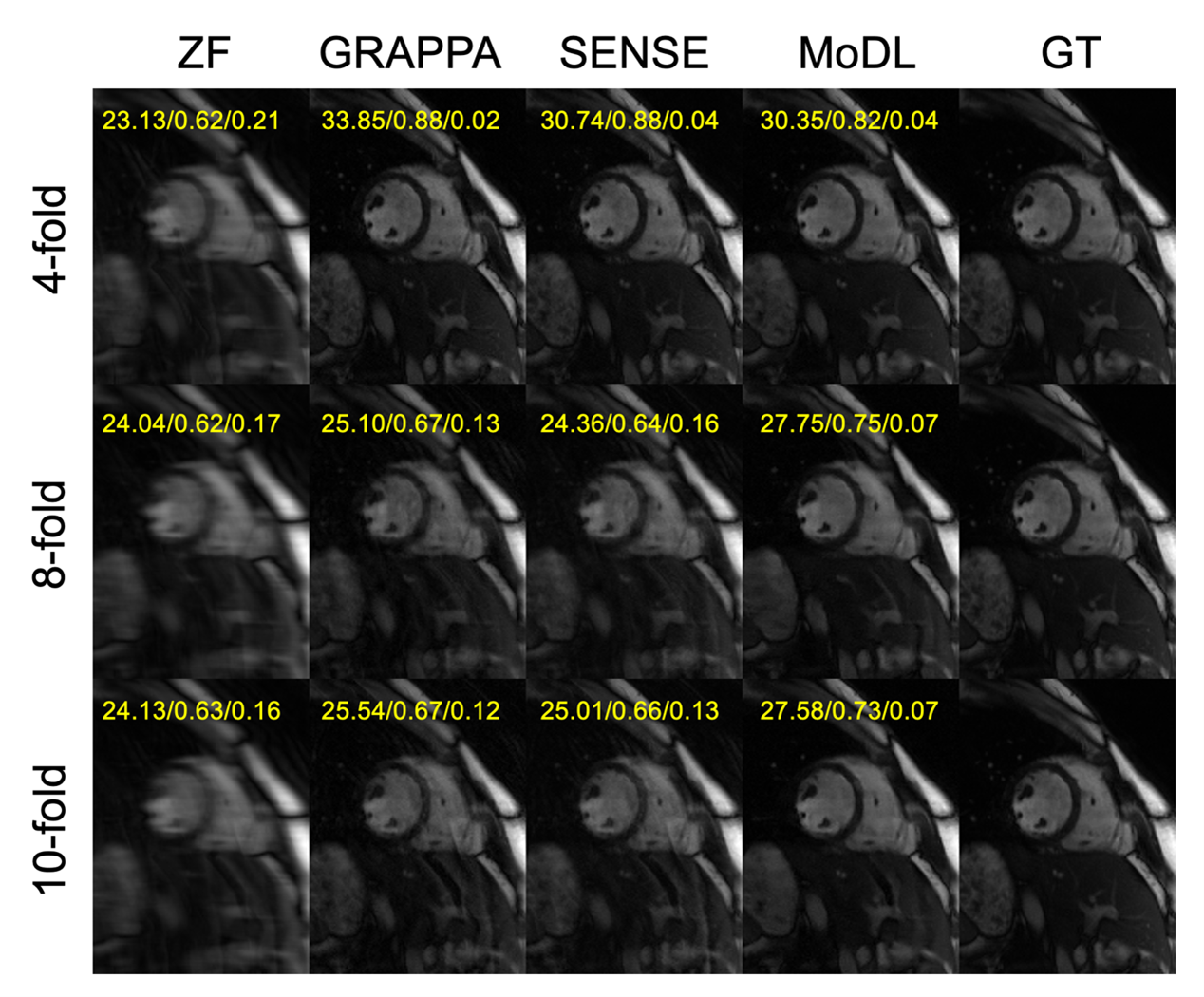}
\caption{Representative Cine short-axis images in the `CMR$\times$Recon' dataset reconstructed with benchmark reconstruction algorithms. The yellow numbers in each subgraph represent PSNR, SSIM and NMSE, respectively.}
\label{fig:recI6}
\end{figure}

\begin{figure}[t!]
\centering
\includegraphics[width=0.8\linewidth]{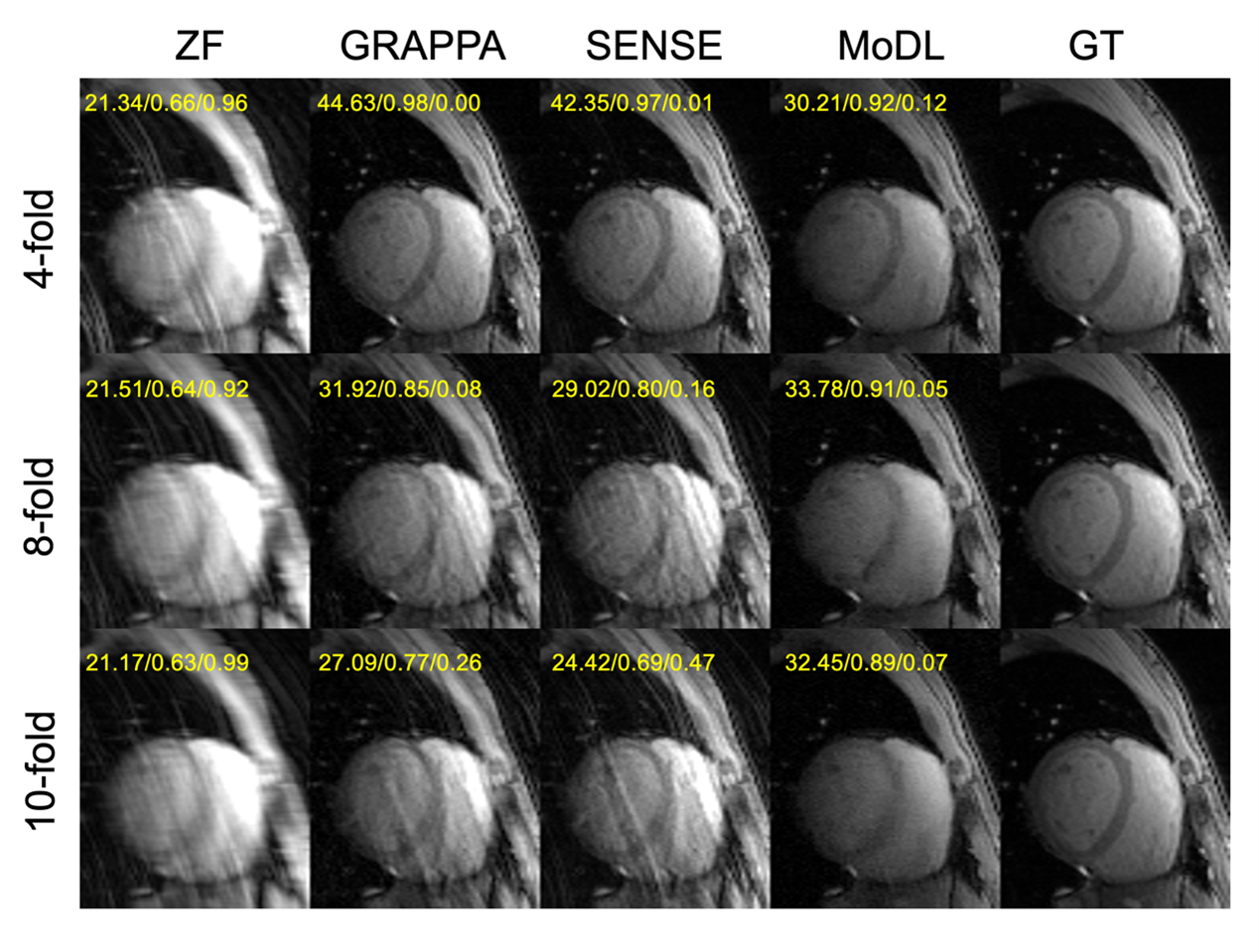}
\caption{Representative T1-mapping images (1st inversion time) in the `CMR$\times$Recon' dataset reconstructed with benchmark reconstruction algorithms. The yellow numbers in each subgraph represent PSNR, SSIM and NMSE, respectively.}
\label{fig:recI7}
\end{figure}

\begin{figure}[t!]
\centering
\includegraphics[width=0.8\linewidth]{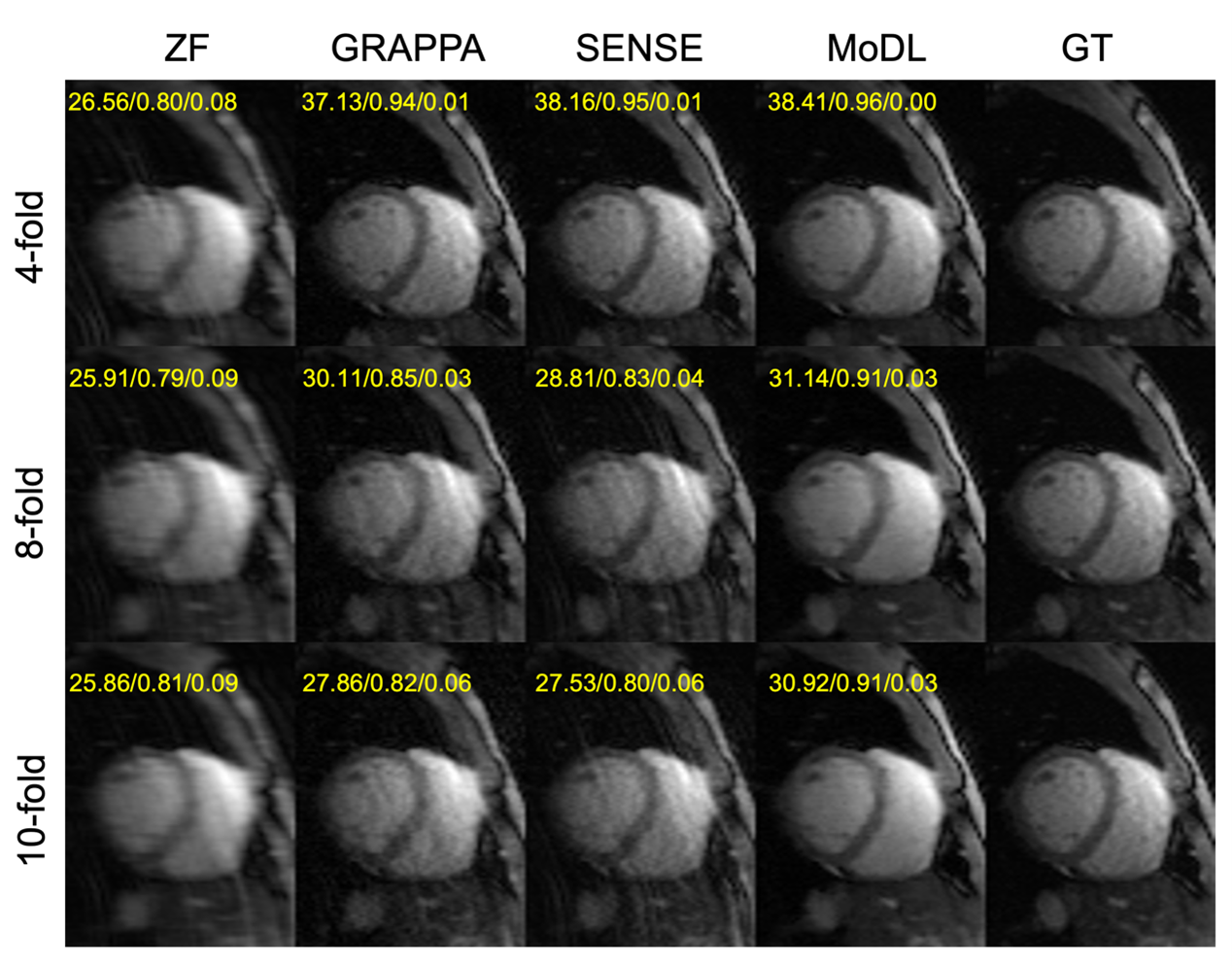}
\caption{Representative T2-mapping images (2nd echo) in the `CMR$\times$Recon' dataset reconstructed with benchmark reconstruction algorithms. The yellow numbers in each subgraph represent PSNR, SSIM and NMSE, respectively.}
\label{fig:recI8}
\end{figure}

We provided scripts to reconstruct the released data using GRAPPA ~\cite{griswold2002generalized}, ESPIRiT ~\cite{uecker2014espirit} and modified MoDL ~\cite{aggarwal2018modl} in the public GitHub repository: \url{https://github.com/CmrxRecon/CMRxRecon}. We used the above state-of-the-art algorithms as a benchmark for advanced AI-based cardiac image reconstruction. Representative results of the `CMR$\times$Recon' dataset with benchmark reconstruction algorithms are displayed in Figure ~\ref{fig:recI5}-~\ref{fig:recI8}. Quantitative assessments of the results using these benchmark algorithms are summarized in Table~\ref{tab:tab5}-~\ref{tab:tab7}. 

\section*{Usage Notes}
The dataset is public and can be downloaded from the Synapse repository through this link (\url{https://www.synapse.org/\#!Synapse:syn51471091/wiki/}). To process the provided k-space data and images, it is recommended to use the tools we provided in the GitHub repository. In addition to the dataset, we also provided a platform for evaluating reconstruction performance (\url{https://www.synapse.org/\#!Synapse:syn51471091/wiki/622170}).

\section*{Code availability}

We provide the script to facilitate the use of the released data at \url{https://github.com/CmrxRecon/CMRxRecon}. A brief description of the provided package is as follows:
\begin{enumerate}
\item[a)] `CMR$\times$ReconDemo’: contains parallel imaging reconstruction code;

\item[b)] `ChallengeDataFormat’: explain the data and the rules for data evaluation;

\item[c)] `Evaluation’: contains image quality evaluation code for validation and testing;

\item[d)] `Download\_Dataset\_Check’: check whether the dataset is completely and rightly downloaded;

\item[e)] `Submission’: contains the structure for challenge submission.

\end{enumerate}




\section*{Author contributions statement}

Chengyan Wang, Jun Lyu, Chen Qin, Shuo Wang, Xiahai Zhuang, Wenjia Bai and Xiaobo Qu designed the research; Xutong Kuang, Mengting Sun and Meng Liu performed the segmentations; Chengyan Wang and Xinyu Zhang performed data anonymization; Xutong Kuang and Yinghua Chu collected data; Chengyan Wang, Fanwen Wang, Yan Li, Jianhua Jin, Ziqiang Xu, Mengting Sun, Meng Liu, He Wang, and Xiaobo Qu analyzed data; Shuo Wang, Kang Wang, Haoran Wang, and Guang Yang performed quality control of the data; Chengyan Wang, Jun Lyu, Jin Qin and Xiaobo Qu wrote the paper; All authors revised and corrected the manuscript. Chengyan Wang and Jun Lyu contributed equally to the paper and Jing Qin, and Xiaobo Qu are senior authors of this manuscript. 

\section*{Competing interests}

None.

\bibliography{sample}

\begin{thebibliography}{10}
\urlstyle{rm}
\expandafter\ifx\csname url\endcsname\relax
  \def\url#1{\texttt{#1}}\fi
\expandafter\ifx\csname urlprefix\endcsname\relax\def\urlprefix{URL }\fi
\expandafter\ifx\csname doiprefix\endcsname\relax\def\doiprefix{DOI: }\fi
\providecommand{\bibinfo}[2]{#2}
\providecommand{\eprint}[2][]{\url{#2}}

\bibitem{nakamura2018mechanisms}
\bibinfo{author}{Nakamura, M.} \& \bibinfo{author}{Sadoshima, J.}
\newblock \bibinfo{journal}{\bibinfo{title}{Mechanisms of physiological and
  pathological cardiac hypertrophy}}.
\newblock {\emph{\JournalTitle{Nature Reviews Cardiology}}}
  \textbf{\bibinfo{volume}{15}}, \bibinfo{pages}{387--407}
  (\bibinfo{year}{2018}).

\bibitem{jaffe2006biomarkers}
\bibinfo{author}{Jaffe, A.~S.}, \bibinfo{author}{Babuin, L.} \&
  \bibinfo{author}{Apple, F.~S.}
\newblock \bibinfo{journal}{\bibinfo{title}{Biomarkers in acute cardiac
  disease: the present and the future}}.
\newblock {\emph{\JournalTitle{Journal of the American college of cardiology}}}
  \textbf{\bibinfo{volume}{48}}, \bibinfo{pages}{1--11} (\bibinfo{year}{2006}).

\bibitem{bai2020population}
\bibinfo{author}{Bai, W.} \emph{et~al.}
\newblock \bibinfo{journal}{\bibinfo{title}{A population-based phenome-wide
  association study of cardiac and aortic structure and function}}.
\newblock {\emph{\JournalTitle{Nature medicine}}}
  \textbf{\bibinfo{volume}{26}}, \bibinfo{pages}{1654--1662}
  (\bibinfo{year}{2020}).

\bibitem{bai2021longitudinal}
\bibinfo{author}{Bai, W.} \emph{et~al.}
\newblock \bibinfo{journal}{\bibinfo{title}{Longitudinal changes of cardiac and
  aortic imaging phenotypes following covid-19 in the uk biobank cohort}}.
\newblock {\emph{\JournalTitle{medRxiv}}} \bibinfo{pages}{2021--11}
  (\bibinfo{year}{2021}).

\bibitem{wang2021recommendation}
\bibinfo{author}{Wang, C.} \emph{et~al.}
\newblock \bibinfo{journal}{\bibinfo{title}{Recommendation for cardiac magnetic
  resonance imaging-based phenotypic study: imaging part}}.
\newblock {\emph{\JournalTitle{Phenomics}}} \textbf{\bibinfo{volume}{1}},
  \bibinfo{pages}{151--170} (\bibinfo{year}{2021}).

\bibitem{wang2019black}
\bibinfo{author}{Wang, C.} \emph{et~al.}
\newblock \bibinfo{journal}{\bibinfo{title}{Black blood myocardial t 2
  mapping}}.
\newblock {\emph{\JournalTitle{Magnetic resonance in medicine}}}
  \textbf{\bibinfo{volume}{81}}, \bibinfo{pages}{153--166}
  (\bibinfo{year}{2019}).

\bibitem{lyu2023region}
\bibinfo{author}{Lyu, J.} \emph{et~al.}
\newblock \bibinfo{journal}{\bibinfo{title}{Region-focused multi-view
  transformer-based generative adversarial network for cardiac cine mri
  reconstruction}}.
\newblock {\emph{\JournalTitle{Medical Image Analysis}}}
  \textbf{\bibinfo{volume}{85}}, \bibinfo{pages}{102760}
  (\bibinfo{year}{2023}).

\bibitem{qin2018convolutional}
\bibinfo{author}{Qin, C.} \emph{et~al.}
\newblock \bibinfo{journal}{\bibinfo{title}{Convolutional recurrent neural
  networks for dynamic mr image reconstruction}}.
\newblock {\emph{\JournalTitle{IEEE transactions on medical imaging}}}
  \textbf{\bibinfo{volume}{38}}, \bibinfo{pages}{280--290}
  (\bibinfo{year}{2018}).

\bibitem{qin2021complementary}
\bibinfo{author}{Qin, C.} \emph{et~al.}
\newblock \bibinfo{journal}{\bibinfo{title}{Complementary time-frequency domain
  networks for dynamic parallel mr image reconstruction}}.
\newblock {\emph{\JournalTitle{Magnetic Resonance in Medicine}}}
  \textbf{\bibinfo{volume}{86}}, \bibinfo{pages}{3274--3291}
  (\bibinfo{year}{2021}).

\bibitem{lv2020parallel}
\bibinfo{author}{Lv, J.}, \bibinfo{author}{Wang, P.}, \bibinfo{author}{Tong,
  X.} \& \bibinfo{author}{Wang, C.}
\newblock \bibinfo{journal}{\bibinfo{title}{Parallel imaging with a combination
  of sensitivity encoding and generative adversarial networks}}.
\newblock {\emph{\JournalTitle{Quantitative Imaging in Medicine and Surgery}}}
  \textbf{\bibinfo{volume}{10}}, \bibinfo{pages}{2260} (\bibinfo{year}{2020}).

\bibitem{lyu2022dudocaf}
\bibinfo{author}{Lyu, J.} \emph{et~al.}
\newblock \bibinfo{title}{Dudocaf: Dual-domain cross-attention fusion with
  recurrent transformer for fast multi-contrast mr imaging}.
\newblock In \emph{\bibinfo{booktitle}{International Conference on Medical
  Image Computing and Computer-Assisted Intervention}},
  \bibinfo{pages}{474--484} (\bibinfo{organization}{Springer},
  \bibinfo{year}{2022}).

\bibitem{wang2022extreme}
\bibinfo{author}{Wang, S.} \emph{et~al.}
\newblock \bibinfo{journal}{\bibinfo{title}{The extreme cardiac mri analysis
  challenge under respiratory motion (cmrxmotion)}}.
\newblock {\emph{\JournalTitle{arXiv preprint arXiv:2210.06385}}}
  (\bibinfo{year}{2022}).

\bibitem{hammernik2018learning}
\bibinfo{author}{Hammernik, K.} \emph{et~al.}
\newblock \bibinfo{journal}{\bibinfo{title}{Learning a variational network for
  reconstruction of accelerated mri data}}.
\newblock {\emph{\JournalTitle{Magnetic resonance in medicine}}}
  \textbf{\bibinfo{volume}{79}}, \bibinfo{pages}{3055--3071}
  (\bibinfo{year}{2018}).

\bibitem{aggarwal2018modl}
\bibinfo{author}{Aggarwal, H.~K.}, \bibinfo{author}{Mani, M.~P.} \&
  \bibinfo{author}{Jacob, M.}
\newblock \bibinfo{journal}{\bibinfo{title}{Modl: Model-based deep learning
  architecture for inverse problems}}.
\newblock {\emph{\JournalTitle{IEEE transactions on medical imaging}}}
  \textbf{\bibinfo{volume}{38}}, \bibinfo{pages}{394--405}
  (\bibinfo{year}{2018}).

\bibitem{knoll2020fastmri}
\bibinfo{author}{Knoll, F.} \emph{et~al.}
\newblock \bibinfo{journal}{\bibinfo{title}{fastmri: A publicly available raw
  k-space and dicom dataset of knee images for accelerated mr image
  reconstruction using machine learning}}.
\newblock {\emph{\JournalTitle{Radiology: Artificial Intelligence}}}
  \textbf{\bibinfo{volume}{2}}, \bibinfo{pages}{e190007}
  (\bibinfo{year}{2020}).

\bibitem{zhao2022fastmri+}
\bibinfo{author}{Zhao, R.} \emph{et~al.}
\newblock \bibinfo{journal}{\bibinfo{title}{fastmri+, clinical pathology
  annotations for knee and brain fully sampled magnetic resonance imaging
  data}}.
\newblock {\emph{\JournalTitle{Scientific Data}}} \textbf{\bibinfo{volume}{9}},
  \bibinfo{pages}{152} (\bibinfo{year}{2022}).

\bibitem{tibrewala2023fastmri}
\bibinfo{author}{Tibrewala, R.} \emph{et~al.}
\newblock \bibinfo{journal}{\bibinfo{title}{Fastmri prostate: A publicly
  available, biparametric mri dataset to advance machine learning for prostate
  cancer imaging}}.
\newblock {\emph{\JournalTitle{arXiv preprint arXiv:2304.09254}}}
  (\bibinfo{year}{2023}).

\bibitem{zhang2013coil}
\bibinfo{author}{Zhang, T.}, \bibinfo{author}{Pauly, J.~M.},
  \bibinfo{author}{Vasanawala, S.~S.} \& \bibinfo{author}{Lustig, M.}
\newblock \bibinfo{journal}{\bibinfo{title}{Coil compression for accelerated
  imaging with cartesian sampling}}.
\newblock {\emph{\JournalTitle{Magnetic resonance in medicine}}}
  \textbf{\bibinfo{volume}{69}}, \bibinfo{pages}{571--582}
  (\bibinfo{year}{2013}).

\bibitem{haacke1991fast}
\bibinfo{author}{Haacke, E.~M.}, \bibinfo{author}{Lindskogj, E.} \&
  \bibinfo{author}{Lin, W.}
\newblock \bibinfo{journal}{\bibinfo{title}{A fast, iterative, partial-fourier
  technique capable of local phase recovery}}.
\newblock {\emph{\JournalTitle{Journal of Magnetic Resonance (1969)}}}
  \textbf{\bibinfo{volume}{92}}, \bibinfo{pages}{126--145}
  (\bibinfo{year}{1991}).

\bibitem{tygert2020simulating}
\bibinfo{author}{Tygert, M.} \& \bibinfo{author}{Zbontar, J.}
\newblock \bibinfo{journal}{\bibinfo{title}{Simulating single-coil mri from the
  responses of multiple coils}}.
\newblock {\emph{\JournalTitle{Communications in Applied Mathematics and
  Computational Science}}} \textbf{\bibinfo{volume}{15}},
  \bibinfo{pages}{115--127} (\bibinfo{year}{2020}).

\bibitem{griswold2002generalized}
\bibinfo{author}{Griswold, M.~A.} \emph{et~al.}
\newblock \bibinfo{journal}{\bibinfo{title}{Generalized autocalibrating
  partially parallel acquisitions (grappa)}}.
\newblock {\emph{\JournalTitle{Magnetic Resonance in Medicine: An Official
  Journal of the International Society for Magnetic Resonance in Medicine}}}
  \textbf{\bibinfo{volume}{47}}, \bibinfo{pages}{1202--1210}
  (\bibinfo{year}{2002}).

\bibitem{uecker2014espirit}
\bibinfo{author}{Uecker, M.} \emph{et~al.}
\newblock \bibinfo{journal}{\bibinfo{title}{Espirit—an eigenvalue approach to
  autocalibrating parallel mri: where sense meets grappa}}.
\newblock {\emph{\JournalTitle{Magnetic resonance in medicine}}}
  \textbf{\bibinfo{volume}{71}}, \bibinfo{pages}{990--1001}
  (\bibinfo{year}{2014}).

\end{thebibliography}

\section*{Figures \& Tables}

\begin{table}[]

\centering
\begin{threeparttable} 
\caption{Scan parameters for the cardiac imaging protocol.}
\label{tab:tab1}
\begin{tabular}{|l|l|l|l|l|}
\hline
\textbf{Parameter}                        & \textbf{Cine   LAX} & \textbf{Cine   SAX} & \textbf{T1   mapping} & \textbf{T2   mapping} \\ \hline
\textbf{TR   (ms)}               & 3.6                 & 3.6                 & 2.67                  & 3.06                  \\ \hline
\textbf{TE (ms)}                 & 1.6                 & 1.6                 & 1.13                  & 1.29                  \\ \hline
\textbf{Flip   angle (°)}        & 48                  & 44                  & 35                    & 12                    \\ \hline
\textbf{Spatial resolution (mm)} & 1.5 $\times$ 1.5           & 1.5 $\times$ 1.5           & 1.5 $\times$ 1.5             & 1.5 $\times$ 1.5             \\ \hline
\textbf{Slice   thickness (mm)}  & 6.0                 & 8.0                 & 5.0                   & 5.0                   \\ \hline
\textbf{FOV (mm)}                & 340 $\times$ 300           & 340 $\times$ 340           & 340 $\times$ 340             & 340 $\times$ 340             \\ \hline
\textbf{Number   of slices}      & 1                   & 11                  & 6                     & 6                     \\ \hline
\textbf{GRAPPA factor*}          & 3.0                 & 3.0                 & 2.0                   & 2.0                   \\ \hline
\textbf{Breath   holds}          & 2                   & 11                  & 6                     & 6                     \\ \hline

\end{tabular}
      \begin{tablenotes} 
		\item GRAPPA = GeneRalized Autocalibrating Partially Parallel Acquisition; FOV = field-of-view; LAX = long-axis; SAX = short-axis; TE = echo time; TR = repetition time.
*The GRAPPA factor is a parameter used in parallel imaging techniques for accelerating MRI acquisition, which represents the level of under-sampling in k-space. 
     \end{tablenotes} 
\end{threeparttable} 
\end{table}

\begin{table}[]
\centering
\caption{Overview of the `CMR$\times$Recon’ dataset.}
\label{tab:tab2}
\begin{tabular}{|c|c|c|c|c|c|}
\hline
\textbf{Modality}                 & \textbf{Dataset} & \textbf{Num of Case} & \textbf{Num of Volume} & \textbf{Num of Slice} & \textbf{Num of Time frame} \\ \hline
\multirow{3}{*}{\textbf{Cine}}    & Training set     & 120                  & 222                    & 1492                  & 2664                       \\ \cline{2-6} 
                                  & Validation set   & 60                   & 111                    & 742                   & 1332                       \\ \cline{2-6} 
                                  & Test set         & 94                   & 158                    & 1090                  & 1896                       \\ \hline
\multirow{3}{*}{\textbf{Mapping}} & Training set     & 120                  & 240                    & 1298                  & 1440                       \\ \cline{2-6} 
                                  & Validation set   & 59                   & 118                    & 634                   & 708                        \\ \cline{2-6} 
                                  & Test set         & 108                  & 158                    & 1162                  & 1293                       \\ \hline
\end{tabular}
\end{table}

\begin{table}[]
\centering

\caption{Details of the data types of cardiac cine. }
\label{tab:tab3}
\scalebox{0.9}{
\begin{threeparttable} 
\begin{tabular}{|l|l|l|l|}
\hline
\textbf{Coil number}                   & \textbf{File/Folder name}  & \textbf{Dimension} & \textbf{Description}                                                               \\ \hline
\multirow{6}{*}{\textbf{Multi-coil}}   & cine\_lax.mat              & (kx,ky,sc,sz,t)    & Complex k-space data of cine with long axis view                                   \\ \cline{2-4} 
                                       & cine\_lax\_mask.mat        & (kx,ky)            & Subsampling mask with long axis view                                               \\ \cline{2-4} 
                                       & cine\_lax                  & -                  & Snapshot images for different slices (start with s) and time frames (start with t) \\ \cline{2-4} 
                                       & cine\_sax.mat              & (kx,ky,sc,sz,t)    & Complex k-space data of cine with short axis view                                  \\ \cline{2-4} 
                                       & cine\_sax\_mask.mat        & (kx,ky)            & Subsampling mask with short axis view                                              \\ \cline{2-4} 
                                       & cine\_sax                  & -                  & Snapshot images for different slices (start with s) and time frames (start with t) \\ \hline
\multirow{10}{*}{\textbf{Single-coil}} & cine\_lax.mat              & (kx,ky,sz,t)       & Complex k-space data of cine with long axis view                                   \\ \cline{2-4} 
                                       & cine\_lax\_mask.mat        & (kx,ky)            & Subsampling mask with long axis view                                               \\ \cline{2-4} 
                                       & cine\_lax                  & -                  & Snapshot images for different slices (start with s) and time frames (start with t) \\ \cline{2-4} 
                                       & cine\_lax\_forlabel.nii.gz & (sx,sy,sz)         & Reconstructed images in NIFTI format (for reference of manual annotation)          \\ \cline{2-4} 
                                       & cine\_lax\_label.nii.gz    & (sx,sy,sz)         & Manual annotated labels in NIFTI format                                            \\ \cline{2-4} 
                                       & cine\_sax.mat              & (kx,ky,sz,t)       & Complex k-space data of cine with short axis view                                  \\ \cline{2-4} 
                                       & cine\_sax\_mask.mat        & (kx,ky)            & Subsampling mask with short axis view                                              \\ \cline{2-4} 
                                       & cine\_sax                  & -                  & Snapshot images for different slices (start with s) and time frames (start with t) \\ \cline{2-4} 
                                       & cine\_sax\_forlabel.nii.gz & (sx,sy,sz)         & Reconstructed images in NIFTI format (for reference of manual annotation)          \\ \cline{2-4} 
                                       & cine\_sax\_label.nii.gz    & (sx,sy,sz)         & Manual annotated labels in NIFTI format                                            \\ \hline
\end{tabular}
      \begin{tablenotes} 
		\item Note. kx: matrix size in x-axis (k-space); ky: matrix size in y-axis (k-space); sc: coil array number (compressed to 10); sx: matrix size in x-axis (image); sy: matrix size in y-axis (image); sz: slice number for short axis view, or slice group for long axis (i.e., 3ch, 2ch and 4ch views); t: time frame. 
     \end{tablenotes} 
\end{threeparttable} 
}

\end{table}

\begin{table}[]
\centering
\caption{Details of the data types of T1 mapping and T2 mapping.}
\label{tab:tab4}
\scalebox{0.9}{
\begin{threeparttable} 
\begin{tabular}{|l|l|l|l|}
\hline
\textbf{Coil number}                   & \textbf{File/Folder name} & \textbf{Dimension} & \textbf{Description}                                                             \\ \hline
\multirow{8}{*}{\textbf{Multi-coil}}   & T1map.mat                 & (kx,ky,sc,sz,w)    & Complex k-space data of T1 mapping                                               \\ \cline{2-4} 
                                       & T1map\_mask.mat           & (kx,ky)            & Subsampling k-space mask                                                         \\ \cline{2-4} 
                                       & T1map.csv                 & -                  & Inversion time for MOLLI (ms)                                                    \\ \cline{2-4} 
                                       & T1map                     & -                  & Snapshot images for different slices (start with s) and weighting (start with t) \\ \cline{2-4} 
                                       & T2map.mat                 & (kx,ky,sc,sz,w)    & Complex k-space data of T2 mapping                                               \\ \cline{2-4} 
                                       & T2map\_mask.mat           & (kx,ky)            & Subsampling k-space mask                                                         \\ \cline{2-4} 
                                       & T2map.csv                 & -                  & Echo time for T2prep-SSFP (ms)                                                   \\ \cline{2-4} 
                                       & T2map                     & -                  & Snapshot images for different slices (start with s) and weighting (start with t) \\ \hline
\multirow{12}{*}{\textbf{Single-coil}} & T1map.mat                 & (kx,ky,sz,w)       & Complex k-space data of T1 mapping                                               \\ \cline{2-4} 
                                       & T1map\_mask.mat           & (kx,ky)            & Subsampling k-space mask                                                         \\ \cline{2-4} 
                                       & T1map.csv                 & -                  & Inversion time for MOLLI (ms)                                                    \\ \cline{2-4} 
                                       & T1map                     & -                  & Snapshot images for different slices (start with s) and weighting (start with t) \\ \cline{2-4} 
                                       & T1map\_forlabel.nii.gz    & (sx,sy,sz)         & Reconstructed images in NIFTI format (for reference of manual annotation)        \\ \cline{2-4} 
                                       & T1map\_label.nii.gz       & (sx,sy,sz)         & Manual annotated labels in NIFTI format                                          \\ \cline{2-4} 
                                       & T2map.mat                 & (kx,ky,sz,w)       & Complex k-space data of T2 mapping                                               \\ \cline{2-4} 
                                       & T2map\_mask.mat           & (kx,ky)            & Subsampling k-space mask                                                         \\ \cline{2-4} 
                                       & T2map.csv                 & -                  & Echo time for T2prep-SSFP (ms)                                                   \\ \cline{2-4} 
                                       & T2map                     & -                  & Snapshot images for different slices (start with s) and weighting (start with t) \\ \cline{2-4} 
                                       & T2map\_forlabel.nii.gz    & (sx,sy,sz)         & Reconstructed images in NIFTI format (for reference of manual annotation)        \\ \cline{2-4} 
                                       & T2map\_label.nii.gz       & (sx,sy,sz)         & Manual annotated labels in NIFTI format                                          \\ \hline
\end{tabular}
      \begin{tablenotes} 
		\item Note. kx: matrix size in x-axis (k-space); ky: matrix size in y-axis (k-space); sc: coil array number (compressed to 10); sx: matrix size in x-axis (image); sy: matrix size in y-axis (image); sz: slice number for short axis view; w: number of weighted images. 
     \end{tablenotes} 
\end{threeparttable} 
}
\end{table}

\begin{table}[]
\centering
\begin{threeparttable} 
\caption{Quantitative assessments (PSNR) of the results in validation set using the benchmark reconstruction algorithms.}
\label{tab:tab5}
\begin{tabular}{|c|c|c|cc|cccc|}
\hline
\multirow{2}{*}{\textbf{Modality}} & \multirow{2}{*}{\textbf{Sequence}} & \multirow{2}{*}{\textbf{Acceleration factor}} & \multicolumn{2}{c|}{\textbf{Single Coil}} & \multicolumn{4}{c|}{\textbf{Multi Coil}}                                                             \\ \cline{4-9} 
                                   &                                    &                                               & \multicolumn{1}{c|}{ZF}        & MoDL     & \multicolumn{1}{c|}{ZF}      & \multicolumn{1}{c|}{GRAPPA}  & \multicolumn{1}{c|}{SENSE}   & MoDL    \\ \hline
\multirow{6}{*}{\textbf{Cine}}     & \multirow{3}{*}{Lax}               & 4-fold                                        & \multicolumn{1}{c|}{22.1516}   & 28.7589  & \multicolumn{1}{c|}{22.7541} & \multicolumn{1}{c|}{33.7013} & \multicolumn{1}{c|}{31.5708} & 33.0039 \\ \cline{3-9} 
                                   &                                    & 8-fold                                        & \multicolumn{1}{c|}{22.4205}   & 26.9347  & \multicolumn{1}{c|}{22.9013} & \multicolumn{1}{c|}{24.7654} & \multicolumn{1}{c|}{24.8402} & 30.5123 \\ \cline{3-9} 
                                   &                                    & 10-fold                                       & \multicolumn{1}{c|}{22.2378}   & 26.1755  & \multicolumn{1}{c|}{22.7744} & \multicolumn{1}{c|}{24.6867} & \multicolumn{1}{c|}{24.6165} & 28.6928 \\ \cline{2-9} 
                                   & \multirow{3}{*}{Sax}               & 4-fold                                        & \multicolumn{1}{c|}{23.9300}   & 29.6782  & \multicolumn{1}{c|}{24.4256} & \multicolumn{1}{c|}{37.4157} & \multicolumn{1}{c|}{36.5195} & 36.8019 \\ \cline{3-9} 
                                   &                                    & 8-fold                                        & \multicolumn{1}{c|}{23.3555}   & 27.9196  & \multicolumn{1}{c|}{23.7012} & \multicolumn{1}{c|}{27.8742} & \multicolumn{1}{c|}{28.6576} & 32.1232 \\ \cline{3-9} 
                                   &                                    & 10-fold                                       & \multicolumn{1}{c|}{23.1276}   & 27.6946  & \multicolumn{1}{c|}{23.4018} & \multicolumn{1}{c|}{26.7013} & \multicolumn{1}{c|}{27.2887} & 30.9681 \\ \hline
\multirow{6}{*}{\textbf{Mapping}}  & \multirow{3}{*}{T1 mapping}        & 4-fold                                        & \multicolumn{1}{c|}{22.7571}   & 30.2547  & \multicolumn{1}{c|}{22.9962} & \multicolumn{1}{c|}{38.5311} & \multicolumn{1}{c|}{38.9250} & 37.8136 \\ \cline{3-9} 
                                   &                                    & 8-fold                                        & \multicolumn{1}{c|}{22.1520}   & 27.8719  & \multicolumn{1}{c|}{22.3764} & \multicolumn{1}{c|}{27.1603} & \multicolumn{1}{c|}{28.0209} & 33.0394 \\ \cline{3-9} 
                                   &                                    & 10-fold                                       & \multicolumn{1}{c|}{22.0235}   & 27.6281  & \multicolumn{1}{c|}{22.2313} & \multicolumn{1}{c|}{24.7443} & \multicolumn{1}{c|}{25.8830} & 30.5629 \\ \cline{2-9} 
                                   & \multirow{3}{*}{T2 mapping}        & 4-fold                                        & \multicolumn{1}{c|}{23.1876}   & 30.0251  & \multicolumn{1}{c|}{23.8863} & \multicolumn{1}{c|}{36.3358} & \multicolumn{1}{c|}{34.8067} & 36.4724 \\ \cline{3-9} 
                                   &                                    & 8-fold                                        & \multicolumn{1}{c|}{22.8897}   & 28.3433  & \multicolumn{1}{c|}{23.3911} & \multicolumn{1}{c|}{27.3059} & \multicolumn{1}{c|}{27.4371} & 32.5929 \\ \cline{3-9} 
                                   &                                    & 10-fold                                       & \multicolumn{1}{c|}{23.0820}   & 28.2114  & \multicolumn{1}{c|}{23.4908} & \multicolumn{1}{c|}{25.9444} & \multicolumn{1}{c|}{26.1902} & 31.6992 \\ \hline
\end{tabular}
      \begin{tablenotes} 
		\item MoDL: model-based deep learning; PSNR: peak signal to noise ratio; ZF: zero-filling. 
     \end{tablenotes} 
\end{threeparttable} 
\end{table}

\begin{table}[]
\centering
\begin{threeparttable} 
\caption{Quantitative assessments (SSIM) of the results in validation set using the benchmark reconstruction algorithms. }
\label{tab:tab6}
\begin{tabular}{|c|c|c|cc|cccc|}
\hline
\multirow{2}{*}{\textbf{Modality}} & \multirow{2}{*}{\textbf{Sequence}} & \multirow{2}{*}{\textbf{Acceleration factor}} & \multicolumn{2}{c|}{\textbf{Single Coil}} & \multicolumn{4}{c|}{\textbf{Multi Coil}}                                                         \\ \cline{4-9} 
                                   &                                    &                                               & \multicolumn{1}{c|}{ZF}        & MoDL     & \multicolumn{1}{c|}{ZF}     & \multicolumn{1}{c|}{GRAPPA} & \multicolumn{1}{c|}{SENSE}  & MoDL   \\ \hline
\multirow{6}{*}{\textbf{Cine}}     & \multirow{3}{*}{Lax}               & 4-fold                                        & \multicolumn{1}{c|}{0.5612}    & 0.8128   & \multicolumn{1}{c|}{0.6400} & \multicolumn{1}{c|}{0.8838} & \multicolumn{1}{c|}{0.8528} & 0.9078 \\ \cline{3-9} 
                                   &                                    & 8-fold                                        & \multicolumn{1}{c|}{0.5707}    & 0.7664   & \multicolumn{1}{c|}{0.6388} & \multicolumn{1}{c|}{0.6842} & \multicolumn{1}{c|}{0.6744} & 0.8582 \\ \cline{3-9} 
                                   &                                    & 10-fold                                       & \multicolumn{1}{c|}{0.5635}    & 0.7382   & \multicolumn{1}{c|}{0.6339} & \multicolumn{1}{c|}{0.6868} & \multicolumn{1}{c|}{0.6704} & 0.8209 \\ \cline{2-9} 
                                   & \multirow{3}{*}{Sax}               & 4-fold                                        & \multicolumn{1}{c|}{0.6443}    & 0.8072   & \multicolumn{1}{c|}{0.7184} & \multicolumn{1}{c|}{0.9305} & \multicolumn{1}{c|}{0.9216} & 0.9334 \\ \cline{3-9} 
                                   &                                    & 8-fold                                        & \multicolumn{1}{c|}{0.6331}    & 0.7650   & \multicolumn{1}{c|}{0.6917} & \multicolumn{1}{c|}{0.7610} & \multicolumn{1}{c|}{0.7715} & 0.8639 \\ \cline{3-9} 
                                   &                                    & 10-fold                                       & \multicolumn{1}{c|}{0.6339}    & 0.7433   & \multicolumn{1}{c|}{0.6848} & \multicolumn{1}{c|}{0.7349} & \multicolumn{1}{c|}{0.7411} & 0.8437 \\ \hline
\multirow{6}{*}{\textbf{Mapping}}  & \multirow{3}{*}{T1 mapping}        & 4-fold                                        & \multicolumn{1}{c|}{0.5898}    & 0.8515   & \multicolumn{1}{c|}{0.6566} & \multicolumn{1}{c|}{0.9447} & \multicolumn{1}{c|}{0.9486} & 0.9435 \\ \cline{3-9} 
                                   &                                    & 8-fold                                        & \multicolumn{1}{c|}{0.5622}    & 0.8095   & \multicolumn{1}{c|}{0.6271} & \multicolumn{1}{c|}{0.7412} & \multicolumn{1}{c|}{0.7768} & 0.8906 \\ \cline{3-9} 
                                   &                                    & 10-fold                                       & \multicolumn{1}{c|}{0.5691}    & 0.8046   & \multicolumn{1}{c|}{0.6300} & \multicolumn{1}{c|}{0.6609} & \multicolumn{1}{c|}{0.7072} & 0.8591 \\ \cline{2-9} 
                                   & \multirow{3}{*}{T2 mapping}        & 4-fold                                        & \multicolumn{1}{c|}{0.6752}    & 0.8656   & \multicolumn{1}{c|}{0.7689} & \multicolumn{1}{c|}{0.9452} & \multicolumn{1}{c|}{0.9220} & 0.9473 \\ \cline{3-9} 
                                   &                                    & 8-fold                                        & \multicolumn{1}{c|}{0.6784}    & 0.8414   & \multicolumn{1}{c|}{0.7524} & \multicolumn{1}{c|}{0.8134} & \multicolumn{1}{c|}{0.8122} & 0.9032 \\ \cline{3-9} 
                                   &                                    & 10-fold                                       & \multicolumn{1}{c|}{0.7073}    & 0.8391   & \multicolumn{1}{c|}{0.7692} & \multicolumn{1}{c|}{0.7874} & \multicolumn{1}{c|}{0.7866} & 0.8958 \\ \hline
\end{tabular}
      \begin{tablenotes} 
		\item MoDL: model-based deep learning; SSIM: structural similarity index measure; ZF: zero-filling. 
     \end{tablenotes} 
\end{threeparttable} 
\end{table}

\begin{table}[htb]
\centering
\begin{threeparttable} 
\caption{Quantitative assessments (NMSE) of the results in the validation set using the benchmark reconstruction algorithms.}
\label{tab:tab7}
\begin{tabular}{|c|c|c|cc|cccc|}
\hline
 &
   &
   &
  \multicolumn{2}{c|}{\textbf{Single Coil}} &
  \multicolumn{4}{c|}{\textbf{Multi Coil}} \\ \cline{4-9} 
\multirow{-2}{*}{\textbf{Modality}} &
  \multirow{-2}{*}{\textbf{Sequence}} &
  \multirow{-2}{*}{\begin{tabular}[c]{@{}c@{}}\textbf{Accelaration  factor}
  \end{tabular}} &
  \multicolumn{1}{c|}{ZF} &
  MoDL &
  \multicolumn{1}{c|}{ZF} &
  \multicolumn{1}{c|}{GRAPPA} &
  \multicolumn{1}{c|}{SENSE} &
  MODL \\ \hline
 &
   &
  4-fold &
  \multicolumn{1}{c|}{ 0.1761} &
    0.0326 &
  \multicolumn{1}{c|}{  0.1656} &
  \multicolumn{1}{c|}{  0.0216} &
  \multicolumn{1}{c|}{  0.0297} &
    0.0166 \\ \cline{3-9} 
 &
   &
  6-fold &
  \multicolumn{1}{c|}{ 0.1764} &
   0.0462 &
  \multicolumn{1}{c|}{ 0.1681} &
  \multicolumn{1}{c|}{ 0.0946} &
  \multicolumn{1}{c|}{ 0.0940} &
   0.0233 \\ \cline{3-9} 
 &
  \multirow{-3}{*}{Lax} &
  8-fold &
  \multicolumn{1}{c|}{ 0.1848} &
  0.0540 &
  \multicolumn{1}{c|}{0.1749} &
  \multicolumn{1}{c|}{0.1024} &
  \multicolumn{1}{c|}{0.1021} &
  0.0365 \\ \cline{2-9} 
 &
   &
  4-fold &
  \multicolumn{1}{c|}{ 0.1448} &
   0.0291 &
  \multicolumn{1}{c|}{ 0.1359} &
  \multicolumn{1}{c|}{ 0.0075} &
  \multicolumn{1}{c|}{ 0.0113} &
   0.0060 \\ \cline{3-9} 
 &
   &
  6-fold &
  \multicolumn{1}{c|}{0.1706} &
  0.0438 &
  \multicolumn{1}{c|}{0.1630} &
  \multicolumn{1}{c|}{0.0543} &
  \multicolumn{1}{c|}{0.0472} &
  0.0164 \\ \cline{3-9} 
\multirow{-6}{*}{\textbf{Cine}} &
  \multirow{-3}{*}{Sax} &
  8-fold &
  \multicolumn{1}{c|}{ 0.1807} &
   0.0478 &
  \multicolumn{1}{c|}{ 0.1743} &
  \multicolumn{1}{c|}{ 0.0741} &
  \multicolumn{1}{c|}{ 0.0648} &
   0.0201 \\ \hline
 &
   &
  4-fold &
  \multicolumn{1}{c|}{0.2205} &
  0.0346 &
  \multicolumn{1}{c|}{0.2151} &
  \multicolumn{1}{c|}{0.0058} &
  \multicolumn{1}{c|}{0.0132} &
  0.0059 \\ \cline{3-9} 
 &
   &
  8-fold &
  \multicolumn{1}{c|}{ 0.2657} &
   0.0615 &
  \multicolumn{1}{c|}{ 0.2576} &
  \multicolumn{1}{c|}{ 0.0730} &
  \multicolumn{1}{c|}{ 0.0657} &
   0.0175 \\ \cline{3-9} 
 &
  \multirow{-3}{*}{\begin{tabular}[c]{@{}c@{}}T1 mapping\end{tabular}} &
  10-fold &
  \multicolumn{1}{c|}{0.2751} &
  0.0621 &
  \multicolumn{1}{c|}{0.2688} &
  \multicolumn{1}{c|}{0.1303} &
  \multicolumn{1}{c|}{0.1006} &
  0.0312 \\ \cline{2-9} 
 &
   &
  4-fold &
  \multicolumn{1}{c|}{ 0.0873} &
   0.0175 &
  \multicolumn{1}{c|}{ 0.0775} &
  \multicolumn{1}{c|}{ 0.0044} &
  \multicolumn{1}{c|}{ 0.0126} &
   0.0041 \\ \cline{3-9} 
 &
   &
  8-fold &
  \multicolumn{1}{c|}{0.0953} &
  0.0259 &
  \multicolumn{1}{c|}{0.0880} &
  \multicolumn{1}{c|}{0.0325} &
  \multicolumn{1}{c|}{0.0388} &
  0.0099 \\ \cline{3-9} 
\multirow{-6}{*}{\textbf{Mapping}} &
  \multirow{-3}{*}{\begin{tabular}[c]{@{}c@{}}T2 mapping\end{tabular}} &
  10-fold &
  \multicolumn{1}{c|}{ 0.0918} &
   0.0267 &
  \multicolumn{1}{c|}{ 0.0863} &
  \multicolumn{1}{c|}{ 0.0447} &
  \multicolumn{1}{c|}{0.0491} &
  0.0121 \\ \hline
\end{tabular}
      \begin{tablenotes} 
		\item MoDL: model-based deep learning; NMSE: normalized mean square error; ZF: zero-filling. 
     \end{tablenotes} 
\end{threeparttable} 
\end{table}






\end{document}